\documentclass[11pt]{article}
\usepackage{graphicx}

\def\markthis#1{{\it #1}}

\begin{document}

\title{A Simple Dynamic Mind-map Framework To Discover Associative Relationships in Transactional Data Streams}
\author{Christoph Schommer\\
         {\small University of Luxembourg, Dept. of Computer Science}\\
         {\small ILIAS Laboratory, MINE Research Group}\\
         {\small 6, Rue Richard Coudenhove-Kalergi, L-1359 Luxembourg, Luxembourg}\\
         {\small Email: christoph.schommer@uni.lu}}
\date{\today}

\maketitle

\begin{abstract}
In this paper, we informally introduce dynamic mind-maps that represent a new approach on the basis of a dynamic construction of connectionist structures during the processing of a data stream. This allows the representation and processing of recursively defined structures and avoids the problem of a more traditional, fixed-size architecture with the processing of input structures of unknown size. For a data stream analysis with association discovery, the incremental analysis of data leads to results on demand. Here, we describe a framework that uses symbolic cells to calculate associations based on transactional data streams as it exists in e.g. bibliographic databases. We follow a natural paradigm of applying simple operations on cells yielding on a mind-map structure that adapts over time. 


\end{abstract}

\section{Introduction}

Mining of Association rules allows finding correlations between items that are grouped into transactions, deducing rules that define relationships between sets of items. This is quite interesting in market basket analysis where the association between products are worth to notice trends or seasonal customer behaviour. Additionally, a support and confidence value is assigned to the discovered rules; they represent the frequency and the strength of the rule, respectively. However, if we take a look at some algorithms to discover association rules, we may observe that associations are still computed when the the input data stream is complete.

\begin{figure}[htbp]
   \centering
   \includegraphics[width=10cm]{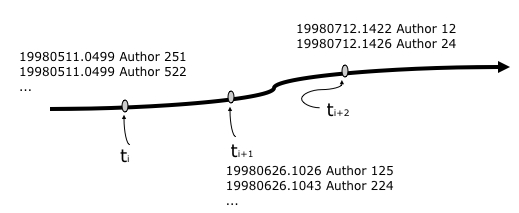} 
   \caption{A transactional data stream.}
   \label{fig:p1}
\end{figure}

There exist many algorithmic strategies as well as sequential and parallel optimisation ideas to overcome complexity problems, to speed up algorithmic time and/or to offer additional supporting parameters, a dynamic and therefore incremental computation, however, is less concerned. In this paper, we therefore introduce a simple dynamic and incremental mind-map that processes incoming transactional data to produce temporal associative relationships. The framework bases on a connectionist network architecture that contains simple symbolic cells and forces natural operations like \markthis{spreading} neural activation, \markthis{merging} of cells, or generation of \markthis{connection weights} to finally come to stable neural architecture.  As for example, Figure \ref{fig:p1} presents a transactional data stream that indicates the publications of authors inside a bibliographic database like \textit{DBLP}, \textit{ACM}, or \textit{coRR}. The assigned stream consists of the transaction identification (including date and reference number), and the name. Two authors share the same publication if the transaction identification is identical. There is no differentiation between author and co-author.

\section{Static approaches}

Designing and implementing algorithms to find associative relationships in a fixed amount of structured data has a long tradition. One of the first algorithms, the \markthis{apriori} algorithm (\cite{AGR93}) bases on simple statistical calculations of \markthis{frequent item sets} computation and association rules generation by counting item frequencies and/or applying Bayesian probabilities. Furthermore, introducing threshold parameters where those are pruned that do not satisfy the threshold solves the complexity problem of having too many item sets. For example, the \markthis{apriori algorithm} (\cite{AGR93}; Figure 1) is based on the calculation of frequent item sets and the generation of association rules by applying statistical operations like counting item frequencies (support value) and Bayesian probability (confidence; the strength of a rule). These statistical values are calculated when the input data stream terminated, i.e. when a full data scan has taken place. Here, \markthis{candidate item sets} that satisfy the user-specified minimum support form the \markthis{frequent item set} whereas \markthis{candidate item sets} that do not reach the threshold are pruned. The trade-off between \markthis{join} and \markthis{prune} reduces the amount of possible candidates.

\begin{figure}[htbp]
   \centering
   \includegraphics[width=14cm]{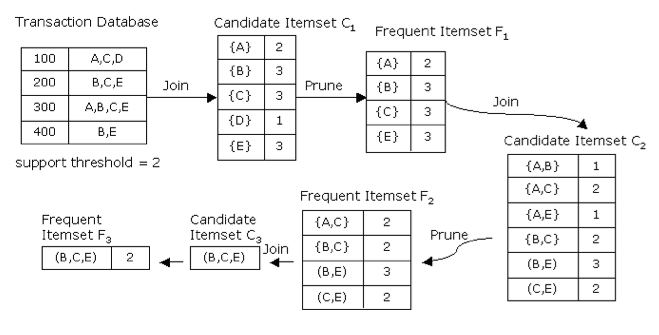} 
   \caption{How the \markthis{apriori} algorithm works for a user-specified support threshold of 2: the discovery process starts if all the data are yet known (\cite{AGR96})}
   \label{fig:p2}
\end{figure}

The \markthis{AprioriTid} algorithm (\cite{THO98}) uses this algorithm to determine the candidate item sets before each pass begins. It does not calculate the frequency after the first pass but uses a set $C_k$ to form a list (TID, $X_k$) where $X_k$ is the potentially large k-item sets present in the transaction with an identifier TID. Based on the above observations, the \markthis{Apriori} Hybrid (\cite{THO98}) uses \markthis{apriori} for the earlier passes and switches to the \markthis{AprioriTid} when the size of the candidate item sets $C_k$ becomes small enough to fit in memory. The partition algorithm (\cite{SHE00}) differs from the \markthis{Apriori} algorithm in terms of the number of full data scans (at most twice). The algorithm is inherently parallel in nature and can be performed in parallel with minimal communication and synchronisation between the processing nodes. The algorithm divides the data into $m$ non-overlapping partitions and generates the frequent item sets for each partition. Local large item sets are merged to form the global candidate item sets; a second data scan is made to generate the final counts.  Set Oriented Mining (\cite{HAN96}) uses join operations (SQL) to generate candidates item set $C_k$. The candidate item set as well the TID of the generating data stream is stored as a sequential structure and is used to count the frequency. There are quite some incremental mining algorithms (\cite{FEL97}, \cite{THU00}) that pick up the aspect of incremental frequent item set discovery with a minimal computation when new transactions are added to or deleted from the transaction database. \cite{THU00} uses a negative border concept (introduced in \cite{TOI96}), which consists of all these item sets that were candidates without having reached the user-specified minimum support. During each pass of the apriori algorithm, the set of candidate item sets $C_k$ is computed from the frequent item sets $F_{k-1}$ in the join and prune phases. The negative border is the set of all those item sets that were candidates in the $k_{th}$ pass but did not satisfy the user specified support, i.e. NB$_d$($F_k$) = $C_k$ - $F_k$. However, the algorithm uses a full scan of the whole database only if the negative border of the frequent item sets expands.

\section{Dynamic Symbolic Cells}

As an alternative to recurrent systems, \markthis{dynamic symbolic cells} are an approach within the connectionist paradigm that is based on the dynamic construction of connectionist structures during the processing of an data input. This allows the representation and processing of recursively defined structures and avoids the problems of a traditional, fixed-size neural network architecture with the processing of input streams of unknown size, for example in transaction-based systems. A longer history of dynamic symbolic cells can be found in parsing natural language and the calculation of the context (\cite{KEM93}, \cite{KEM01}).

Dynamic symbolic cells follow the idea of having \markthis{mini-mind-maps} where these mind-maps are successively composed into larger ones by generating and merging processes. As for example in natural language parsing, these processes are triggered by the occurrence of external inputs, or fully activated nodes within a generated network. Due to the generative character, the processing and representation of inputs is not restricted by a fixed-size but reflects the generative and constructive paradigm, as it is inherent to grammar-based descriptions of natural languages.

Symbolic cells do contain symbolic value that can for example be an item within a transaction database. As regular dynamic cells, symbolic cells are interconnected; they stimulate or inhibit each other. The activation values range from -1.0 (totally inhibiting) to 1.0 (totally stimulating); the network forces \markthis{Hebbian Learning} between activated symbolic cells to update the weights between associated cells. Symbolic cells work highly parallel; the corresponding activation status of each cells as well the connection weights is synchronously updated after an input stream
was read. By nature, symbolic cells may compete in a \markthis{winner-take-all} strategy (\cite{KEM94}, \cite{KEM01}); in this approach, however, they only cooperate in a sense that symbolic cells, which are not activated by input cell, remain untouched. For the Association Discovery, we use symbolic cells as \markthis{Input Cells} and \markthis{Item Cells}: Input cells are assigned to an data stream entry, for example, elements of a single transaction. However, if there are item cells that share the same content (for example, the input stream is 'A,C,D,A'), then these input items get merged by increasing the internal activation status. Item cells derive from the input cells by \markthis{cell division} and establishing weighted connections. Depending on the input data stream, input cells occur always when an input is read whereas item cells are created only then, when no item cell exists that shares the same symbolic input. If there exists already a symbolic  cell with this input, the input cell is \markthis{merged} with the existing item cell.

The information which items are associated is stored both in the connection weights between item cells and in the item cells itself. Therefore, the representation of the information management is therefore hybrid; and the discovery of associations is finally be done by simple operations: using \markthis{cell division} and \markthis{merging} symbolic cells can perform efficient communication and information exchange.

Cell division is performed by the input cell; it then establishes undirected connections between all new resulting item cells whereas the (old) input cells die. Cell division allows reducing the number of symbolic cells in a sense that the number of item cells does not increase the number of existing different items in the data stream.

\begin{figure}[htbp]
\begin{center}
  \includegraphics[width=14cm]{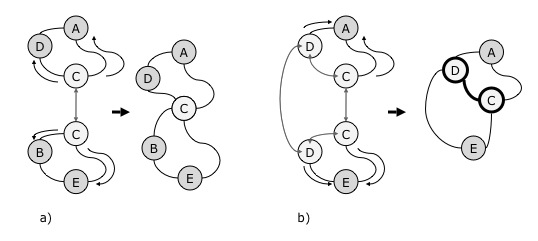}
  \label{fig:p4}
  \caption{ Merging - spreading out activation and changing the connection weights by \markthis{Hebbian Learning}: if both cells are active then the connection weight between symbolic cells increases (b), otherwise not (a).}
 \end{center}
 \end{figure}

\begin{figure}[htbp]
\begin{center}
  \includegraphics[width=10cm]{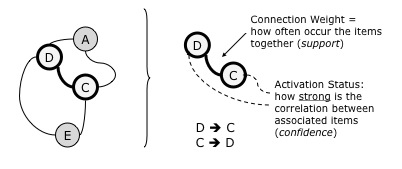}
  \label{fig:p5}
  \caption{At given time point $t_i$ we can find interacting the \markthis{skeleton} of the mind-maps by regarding item cells.}
 \end{center}
 \end{figure}

The process of merging (Figure \ref{fig:p4}) input cells or item cells can only be done in case that the corresponding symbolic cells have the same content. Here, the corresponding item cells stimulate each other by sending out signals (coming from the activation function) that is scaled by their existing connection weight.  All neighbour cells of a merged cell get activated if and only if they are in a merged status. The connection weight between these cell increases (\markthis{Hebbian Learning}). If an item cell does not become activated for a longer time by other item cells in the network (i.e. the corresponding item does not occur in the input stream or is rather seldom), then this item can be \markthis{forgotten} by the network. This process of information loss is similar to the natural behaviour in life science; it will be simulated by using a \markthis{bias} parameter that reduces the connection weight to neighbour cells at synchronization time. After each synchronization step, association rules can be extracted from the existing mind-maps (Figure \ref{fig:p5}).

Depending on user-specified thresholds for connection weights and/or activation status,  the \markthis{skeleton} of the mind-maps can be derived. A simple way to achieve this is to store the contents and the corresponding connection weights within a dynamic hash table matrix of size $n\times n$ having n as the number of item cells.
 
In order to recognise and manage associations, we need a way to store strong patterns along with some additional information on if the pattern has been observed before and when this happened. We therefore have extended the architecture with memory layers. The mind-map is then composed of a short-term and long-term memory that are used to manage the patterns and to maintain strong patterns: these are copied into the long-term memory along with information on when it appeared and when it disappeared again. Appearing strong patterns can then be matched against existing patterns in the long term memory to establish their recurrence. In this respect, the short-term working memory is available for the establishment and maintenance of the association network. A second task is the search for emerging association patterns, respectively, strongly connected subgraphs. If such a pattern is found, it is transferred to the long-term memory and stored. A query and reporting interface bridges the communication. The system informs the user autonomously about its current status and reports certain events, for instance newly discovered patterns. In addition to the typical static queries known from traditional database systems, the user may pose continuous queries which are evaluated over a period of time, for example towards the strongest sub-graphs (following the status of the connection weights). The answer to this query is a one-time output after a certain time-step. The answer to a query can however be a stream as well, for example to trace the weight of connection c over $k$ time-steps.  The queries are targeted at both the short-term and long-term memory.

\begin{figure}[htbp]
   \centering
   \includegraphics[width=10cm]{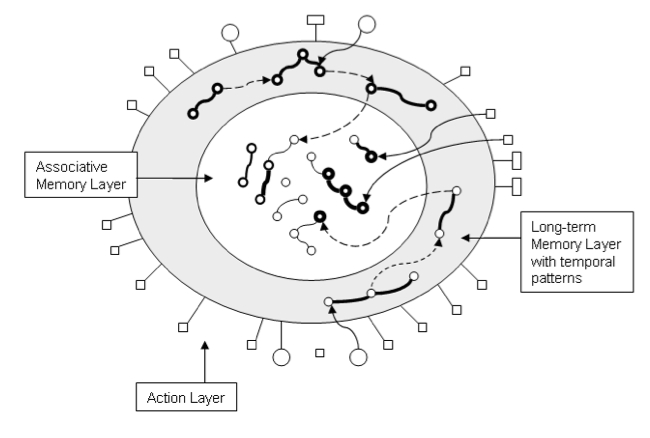} 
   \caption{Introducing a short-term and a long-term memory layer: once a pattern survives over a period of time, it is copied into the long-term memory (from \cite{SCR05b}).}
   \label{fig:p0}
\end{figure}

\section{An Example}

Assume, there is a transactional data stream. Then, the computation of association rules is performed as follows: first, the input \textit{A, A, C, D} is read and four input cells are generated. Because there exist two cells hosting the content \textit{A}, they merge and form a new input cell. Finally, cell division forces the cells to establish connections with an initial connection weight of $\frac{1}{3}$. The item cells form a mind-map having the same activation status.

\begin{figure}[htbp]
\begin{center}
  \includegraphics[width=14cm]{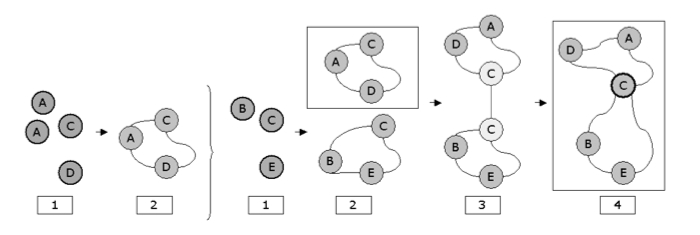}
  \label{fig:p6}
  \caption{Incremental Association Discovery (first and second data stream): reading A,C,D; creating input cells (1) and cell division (2); reading B,C,E; creating input cells (1) and cell division (2); merging the item cells (3, 4)}
 \end{center}
 \end{figure}

The presentation of the second data stream \textit{B, C, E} leads in the first moment to a second mind-map: however, both mind-maps share a common symbolic cell \textit{C} and therefore merge by nature. The activation status of the item cell with content \textit{C} increases; there exist only one mind-map. All input cells have been died.

\begin{figure}[htbp]
\begin{center}
  \includegraphics[width=14cm]{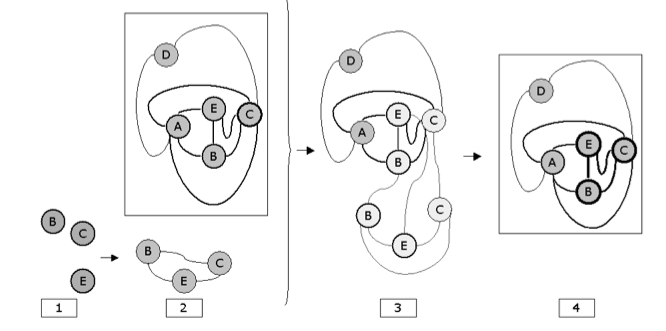}
  \label{fig:p7}
  \caption{ Incremental Association Discovery (fourth data stream): creating input cells (1), cell division (2), and merging (3, 4)}
 \end{center}
 \end{figure}

The presentation of the third data stream \textit{A, B, C, E} leads to a third mind-map; it is obvious that the number of symbolic cells in the mind-maps correspond to the number of disjunctive items in the data stream. However, the activation status of symbolic cells are normally quite different than others depending on the number of occurrences. After presenting of the fourth data stream \textit{B, C, E} and the generation of input and item cells, the merge between item cells sharing the same content within the mind-map takes place. As it is for the \markthis{Hebbian Learning} paradigm, activated symbolic cells learn in a sense that the corresponding connection weights of the associated cells get increased. If a cell is not activated by a merge, however, the connection weight does not increase.

\begin{figure}[htbp]
\begin{center}
  \includegraphics[width=14cm]{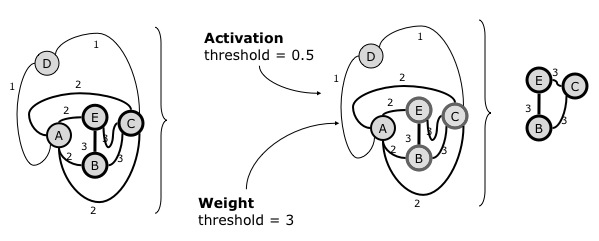}
  \label{fig:p8}
  \caption{ Mind-Map with a higher activated item cells having stronger connection weights than other item cells. The association of items can be noticed in the final skeleton.}
 \end{center}
 \end{figure}

We finally get one mind-map that hosts the corresponding association between the items. By applying a user-specified thresholds for connection weight (here: k)
we then get the skeleton of the mind-map with symbolic cells those connections are strong enough \textit{E, C, B}. The connections of the skeleton are undirected, i.e. both the rules $E \Rightarrow B$, $B \Rightarrow E$, $E \Rightarrow C$, $C \Rightarrow E$, $C \Rightarrow B$, and $B \Rightarrow C$ are associative.

\section{Conclusions}

A disadvantage of the model is that it highly depends on the items' order. For example, if there are millions of disjunctive items where item $n_j$ occurs only twice in the last input streams, then this item will probably have a higher valued connection weights than if it occurred in the first input streams (this item will not become forgotten). Furthermore, the number of item cells remain in the range of $\Omega(n)$ with probably a full connection of $O(n^2)$. Another disadvantage is that associations may appear only pairwise, whereas associative relationships of size $>2$ are not reflected.

The advantages are the basic dynamic behaviour of the model, the scalable architecture, the parallel processing and synchronization character, as well as the tolerance in terms of data noise. Furthermore, the resulting mind-maps - consisting of interacting symbolic cells - can fully be interpreted as a semantic net. For example in market basket analysis, the mind-map can be seen as a real-time snapshot representing the current customers' behavior.

An extension of the existing model is to start with a \markthis{predefined mind-map}
or \markthis{skeleton}, for example as a customer profile in an online-shop where item units represent given websites. This could also be of interest for \markthis{scoring} processes where a given skeleton can allow to suggest most interesting pages. Furthermore, the process of merging could be 'turned around' in a sense that only associations - that occur very seldom - are regarded as interesting: here, the item cells use \markthis{mutual inhibition} in order to decrease the relevance of main stream associations. Item cells that occur very seldom are less inhibited and therefore immediately present. This is quite interesting for the \markthis{detection of anomalies}, for example the \markthis{intrusion detection} problem in a high-traffic network. 

\section{Acknowledgement}

This work has been performed in the MINE Research Group of the Dept. of Computer Science and Communication, University of Luxembourg. It summarizes the fundamental principle of a dynamic mind-map as we it understand. The paper is an updated version of the workshop presentation given in \cite{SCH04a}. Since that time, diverse applications and implementations have been performed; therefore, the author explicitly thanks Claudine Brucks, Yafang Chang, Naipeng Dong, Michael Hilker, Armand Kopczinski, Ben Schroeder, Cynthia Wagner, Ralph Weires, and Ejikem Uzochuwkwu.

\end{document}